\documentclass[pdflatex,sn-mathphys-num]{sn-jnl}

\usepackage{float}
\usepackage{graphicx}%
\usepackage{multirow}%
\usepackage{amsmath,amssymb,amsfonts}%
\usepackage{amsthm}%
\usepackage{mathrsfs}%
\usepackage[title]{appendix}%
\usepackage{xcolor}%
\usepackage{textcomp}%
\usepackage{manyfoot}%
\usepackage{booktabs}%
\usepackage{algorithm}%
\usepackage{algorithmicx}%
\usepackage{algpseudocode}%
\usepackage{listings}%
\usepackage{makecell}   
\usepackage{array}
\usepackage{tabularx}
\usepackage{microtype}
\usepackage[numbers]{natbib}

\theoremstyle{thmstyleone}%
\theoremstyle{thmstyletwo}%

\theoremstyle{thmstylethree}%

\begin{document}

\title[Article Title]{CS-MUNet: A Channel-Spatial Dual-Stream Mamba Network for Multi-Organ Segmentation}

\author[1]{\fnm{Yuyang} \sur{Zheng}}\email{zhengyy@stu.ynu.edu.cn}
\equalcont{These authors contributed equally to this work.}
\author[1,2]{\fnm{Mingda} \sur{Zhang}}\email{mingdazhang@ieee.org}
\equalcont{These authors contributed equally to this work.}
\author*[1,3]{\fnm{Jianglong} \sur{Qin}}\email{qinjianglong@ynu.edu.cn}
\author[1,3]{\fnm{Qi} \sur{Mo}}
\author[1]{\fnm{Jingdan} \sur{Pan}}
\author[1]{\fnm{Haozhe} \sur{Hu}}
\author[1]{\fnm{Hongyi} \sur{Huang}}
\affil[1]{\orgdiv{School of Software Engineering}, \orgname{Yunnan University}, \orgaddress{\city{KunMing}, \postcode{650500}, \state{Yunnan}, \country{China}}}
\affil[2]{\orgdiv{The School of Artificial Intelligence}, \orgname{The Chinese University of Hong Kong}, \orgaddress{\city{Shenzhen}, \postcode{518172}, \state{Guangdong}, \country{China}}}
\affil[3]{\orgdiv{Yunnan Provincial Key Laboratory of Software Engineering}, \orgname{Yunnan University}, \orgaddress{\city{Kunming}, \postcode{650500}, \state{Yunnan}, \country{China}}}


\abstract{Recently Mamba-based methods have shown promise in abdominal organ segmentation.However, existing approaches neglect cross-channel anatomical semantic collaboration and lack explicit boundary-aware feature fusion mechanisms.To address these limitations, we propose CS-MUNet with two purpose-built modules. The Boundary-Aware State Mamba module employs a Bayesian-attention framework to generate pixel-level boundary posterior maps, injected directly into Mamba's core scan parameters to embed boundary awareness into the SSM state transition mechanism, while dual-branch weight allocation enables complementary modulation between global and local structural representations.The Channel Mamba State Aggregation module redefines the channel dimension as the SSM sequence dimension to explicitly model cross-channel anatomical semantic collaboration in a data-driven manner. Experiments on two public benchmarks demonstrate that CS-MUNet consistently outperforms state-of-the-art methods across multiple metrics, establishing a new SSM modeling paradigm that jointly addresses channel semantic collaboration and boundary-aware feature fusion for abdominal multi-organ segmentation.}

\keywords{Mamba,Organ segementation,Feature fusion,Edge-aware}



\maketitle

\section{Introduction}\label{sec1}
\begin{wrapfigure}{r}{0.45\textwidth}
    \vspace{-6pt}
    \centering
    \includegraphics[width=0.45\textwidth]{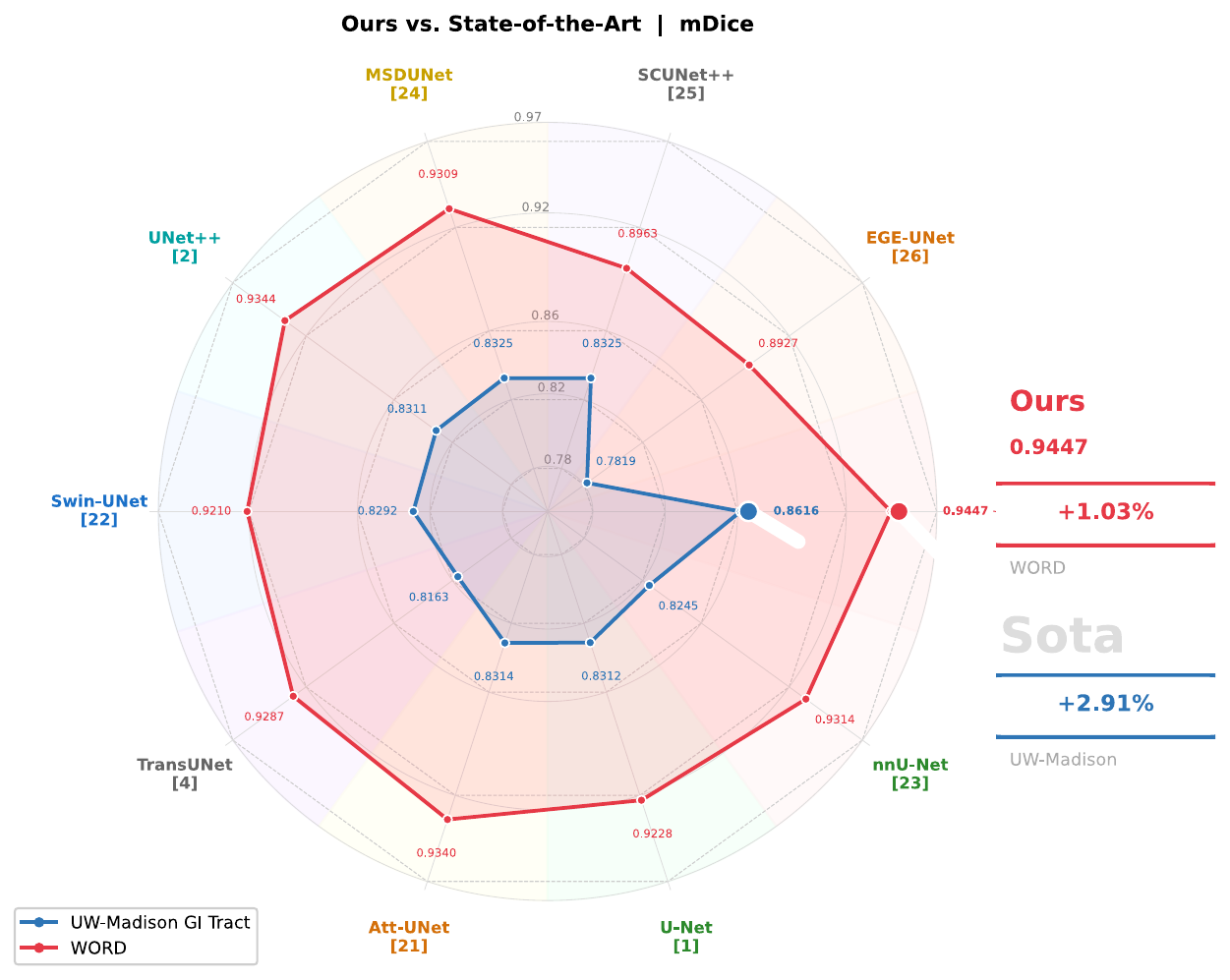}
    \caption{Comparative mDice performance of CS-MUNet against
    state-of-the-art methods on the UW-Madison GI Tract and WORD
    benchmarks. CS-MUNet outperforms all competing methods on both
    datasets, achieving gains of +2.91\% and +1.03\% in mDice over
    the strongest baseline on UW-Madison and WORD respectively,
    demonstrating consistent superiority across both MRI and CT
    modalities.}
    \label{fig:radar}
    \vspace{-8pt}
\end{wrapfigure}
Medical image segmentation is a core task in computer-aided diagnosis. 
Abdominal hollow organs are notoriously difficult to segment due to variable 
morphology, indistinct boundaries, and low soft-tissue contrast in 
CT (Computed Tomography) and MRI (Magnetic Resonance Imaging) 
imaging~\cite{bib26,bib27,bib28,bib29,bib30,bib31}. Existing Mamba-based 
methods universally treat spatial positions as sequence units while neglecting 
cross-channel anatomical semantic collaboration~\cite{bib11,bib12,bib13}, 
and their boundary-aware designs fail to penetrate core parameter modulation 
of state space scanning.\par
Building on U-Net~\cite{bib1}, numerous CNN-based methods~\cite{bib2,bib3,bib4,bib21} 
have improved local feature extraction but are limited by fixed receptive fields. 
Transformers~\cite{bib5,bib6,bib7} enable global modeling at the cost of $O(n^2)$ 
complexity. Mamba~\cite{bib8} and its visual adaptations~\cite{bib9,bib10} achieve 
linear-complexity $O(n)$ long-range modeling, and have been rapidly adopted for 
medical segmentation across U-shaped SSM architectures~\cite{bib11,bib13,bib17,bib32}, 
volumetric segmentation~\cite{bib12,bib16}, lightweight design~\cite{bib14,bib15,bib33}, 
hybrid convolutions~\cite{bib18,bib25,bib34,bib35}, annotation-limited 
settings~\cite{bib36,bib37,bib38}, and clinical applications~\cite{bib39,bib40}.\par
However, three key limitations persist: (1) Channel-dimension semantic collaboration 
is ignored: existing methods~\cite{bib10,bib11,bib12,bib13,bib17,bib32} treat channels 
as independent parallel components, overlooking ordered inter-channel dependencies 
on anatomy-specific semantics such as texture, shape, and density. (2) Boundary 
awareness is not deeply integrated with SSM scanning: while Wang et al.~\cite{bib19} 
demonstrated substantial gains from explicit boundary modeling, existing Mamba 
methods~\cite{bib11,bib13,bib15,bib25} lack boundary-aware modules co-designed with 
SSM parameters, and BAMN~\cite{bib20} only applies boundary constraints as 
post-processing. (3) Hollow-organ-specific optimization is severely lacking: existing 
Mamba methods~\cite{bib12,bib16} are validated primarily on solid organs, ignoring 
thin-wall and weak-boundary challenges of hollow organs~\cite{bib27,bib29}. To this 
end, we propose CS-MUNet, building dual-stream SSM modeling pathways along orthogonal 
channel and spatial dimensions. Our contributions are summarized as follows:
\begin{itemize}
\item We deploy the Boundary-Aware State Mamba module (BASM) at decoder skip 
connections and the Channel Mamba State Aggregation module (CMSA) at the encoder 
bottleneck, addressing spatial boundary awareness and channel semantic 
collaboration respectively.
\item BASM employs a Bayesian-attention framework to generate pixel-level boundary 
posterior maps injected directly into Mamba's core parameters $\Delta$/B, embedding 
boundary awareness into the SSM state transition mechanism itself.
\item CMSA redefines the channel dimension as the SSM sequence dimension to explicitly 
model cross-channel anatomical semantic collaboration, with bounded state transition 
constraints preventing unbounded state accumulation across heterogeneous channels.
\end{itemize}
Extensive experiments on two public benchmarks demonstrate that CS-MUNet consistently 
outperforms existing state-of-the-art methods, validating the effectiveness of the 
proposed architecture.

\section{Related Work}\label{sec2}
\subsection{U-Net and its variants}\label{subsec2}
Deep learning-driven medical image segmentation has evolved from pure convolutional to hybrid architectures. Starting from U-Net\cite{bib1}, researchers have proposed a series of improvements around skip connections, multi-scale fusion, and attention gating\cite{bib2,bib3,bib4,bib21}, yet the local receptive field of convolutions limits global structural modeling. Transformers effectively address this limitation, progressing from CNN-Transformer hybrid encoders\cite{bib5} and pure Transformer U-shaped structures\cite{bib6} to three-dimensional global modeling\cite{bib7}. Subsequently, MSDUNet\cite{bib22} enhances segmentation via multi-scale feature fusion and dual-input dynamic enhancement; SCUNet++\cite{bib23} strengthens multi-scale semantic transfer by combining Swin-UNet with CNN bottleneck structures; EGE-UNet\cite{bib24} achieves competitive boundary segmentation accuracy under low parameter counts through lightweight group enhancement and boundary-aware supervision. Nevertheless, these methods remain constrained by computational complexity.
\subsection{Mamba}\label{subsec2}
Gu and Dao\cite{bib8} proposed Mamba with linear complexity O(n) for long-range modeling, and Zhu et al.\cite{bib9} and Liu et al.\cite{bib10} subsequently adapted it to the visual domain, rapidly driving a wave of medical segmentation research. In architecture design, researchers have progressively established U-shaped segmentation baselines from CNN-SSM hybrid paradigms\cite{bib11} to pure SSM encoder-decoder structures\cite{bib13,bib17,bib32}, and extended to volumetric segmentation through multi-directional scanning and multi-task frameworks\cite{bib12,bib16}. In efficiency optimization, pretraining, parameter compression, and automated scanning strategies\cite{bib14,bib15,bib33,bib35} have pushed model lightweighting to the extreme. In feature enhancement, high-order scanning interactions\cite{bib25}, hybrid convolutions for expanded receptive fields\cite{bib34}, and skip-connection fusion combining boundary-enhanced multi-scale attention with convolutional Mamba\cite{bib41} have effectively improved boundary-aware accuracy. Mamba has further been extended to semi-supervised\cite{bib36}, weakly supervised\cite{bib37}, prompt learning\cite{bib38}, and diverse clinical tasks\cite{bib18,bib39,bib40}. However, existing methods universally neglect cross-channel anatomical semantic collaboration, and boundary information has not penetrated the core parameter modulation of SSM scanning, limiting segmentation accuracy on abdominal hollow organs.
\section{Method}\label{sec3}
\begin{figure*}[htbp]
    \centering
    \includegraphics[width=\textwidth]{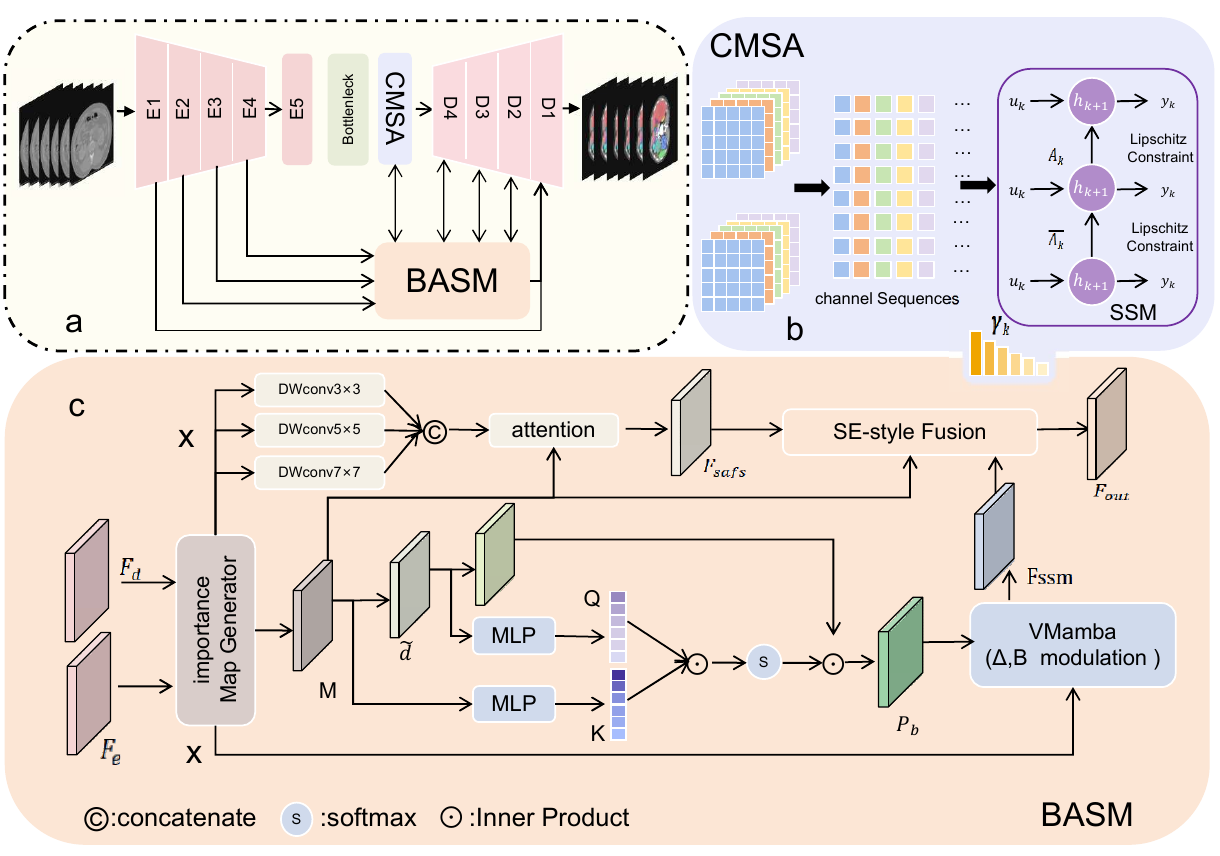}
    \caption{Overall architecture of CS-MUNet and its two proposed modules.(a) U-shaped encoder-decoder with CMSA at the bottleneck and BASM at each skip connection fusing  $F_e$ and decoder feature $F_d$.(b)CMSA processes grouped channel sequences via a shared SSM, where $\bar{A}_k$,$\gamma_k$ denote the state transition matrix and cumulative decay under Lipschitz constraints, and $\mu_k$,$h_{k+1}$, and $y_k$ the input, hidden state, and output of the k-th channel.(c) BASM injects boundary posterior $P_b$ derived from guidance map M and distance field $\tilde{d}$ into VMamba's $\Delta$ and B parameters to yield $F_{ssm}$, fused with $F_{safs}$ to produce $F_{out}$.}
    \label{Figure 2}
\end{figure*}

As illustrated in Figure~\ref{Figure 2}, this section introduces the CS-MUNet architecture, including the overall network framework and components(Section~\ref{Sec3.1}) , the design and implementation of the Boundary-Aware State Mamba module BASM(Section~\ref{Sec3.2}), and the serialized channel modeling mechanism of the Channel Mamba State Aggregation module CMSA(Section~\ref{Sec3.3}).
\subsection{Overall Network Architecture and Components}\label{Sec3.1}
\textbf{Overall Architecture.} Overall Architecture. CS-MUNet adopts Res2Net-50 as backbone, deploying CMSA at the encoder bottleneck for cross-channel semantic recalibration and BASM at each skip connection for boundary-posterior-driven heterogeneous feature fusion, thereby forming an ordered information flow of "channel semantic refinement → spatial boundary-aware fusion."\par
\textbf{Boundary Semantic Guidance Map Construction}. Skip connections feed shallow encoder features $F_e \in \mathbb{R}^{C \times H \times W}$ and deep decoder features $F_d \in \mathbb{R}^{C \times H \times W}$ into BASM. Their pixel-wise alignment exhibits significant spatial non-uniformity—lowest at boundaries and highest in homogeneous interiors—which we explicitly model as the core prior signal of BASM, defining the boundary semantic guidance map $M \in \mathbb{R}^{H \times W}$ as:
\begin{equation}
M(i,j)=\sigma(w_b \cdot Sobel(\frac{F_e(i,j) \cdot F_d(i,j)}{||F_e(i,j)|| \cdot ||F_d(i,j)||})+w_f \cdot P_{fg}(i,j))
\end{equation}
where $w_b$, $w_f$ are learnable scalar weights and $P_{fg}$ is the foreground probability map. M serves as a globally shared prior driving both Mamba parameter modulation and dual-stream weight allocation in BASM, ensuring semantic consistency across both components.\par
\textbf{Spatial Adaptive Structure-aware Feature Branch (SASF)}. Mamba models spatial sequence dependencies via global linear recurrence, which lacks explicit multi-scale inductive bias for local geometric structures. SASF is defined as a local structural prior compensation term for Mamba's global sequence modeling. Given input feature $x \in \mathbb{R}^{C \times H \times W}$ and the shared guidance map M, the SASF output is defined as:
\begin{equation}
F_{safs}=Conv_{1×1}(Concat[DWConvk(x)]_{k\in {3,5,7}}) \odot (1+M)
\end{equation}
where parallel depthwise separable convolutions $\text{DWConv}_k$ capture geometric responses at three scales, and (1+M) acts as a spatial gain factor.\par
\textbf{Proposition 1.} The dual-branch design of BASM, combining global SSM sequence modeling and SASF local geometric compensation, provides complementary boundary reinforcement from two orthogonal dimensions.\par 
\textbf{Proof.} Ablation results are provided in Section~\ref{sec4.3}, Table~\ref{Table3}.
\subsection{Boundary-Aware State Mamba Module (BASM)}\label{Sec3.2}
BASM embeds boundary awareness directly into SSM state transitions by modulating$\Delta$(forgetting rate) and B (write weight), transforming boundary priors from passive post-processing into active scan-guiding signals. It adopts a dual-branch architecture: the Mamba branch executes boundary-posterior-driven parameter modulation, while the SASF branch compensates local perception via multi-scale depthwise convolutions, fused through SE-style weight allocation. \par
\textbf{Formal Motivation.} In the standard selective SSM, the time-step parameter $\Delta$ is generated by linear projection from the input, whose receptive field is confined to the channel vector at the current position and therefore lacks architectural capacity to perceive whether a pixel lies on an organ boundary. Let $\mathcal{B}$ denote the set of boundary pixels and $\mathcal{U}$ the set of background pixels. The standard SSM applies statistically indistinguishable state transitions to both:
\begin{equation}
    E[\Delta(i)\mid i\in \mathcal{B}]=E[\Delta(i)\mid i\in \mathcal{U}]
\end{equation}
This structural deficiency is architecture-determined rather than optimization-solvable. BASM targets precisely this equality, inducing statistically significant differentiated responses in $\Delta$ and B between boundary and background regions.\par
\textbf{Boundary Posterior Generation.} We factorize boundary evidence into two orthogonal terms under a Bayesian framework: geometric prior $\tilde{d}$, encoding global topology via exponential compression of the distance field from binarized M(threshold $\tau$,decay $\alpha$); and attention likelihood L, jointly conditioning on feature semantics and geometry by projecting M and $\tilde{d}$,into Q-K pairs via independent MLPs, temperature-scaled by $\gamma$,$P_b$ is obtained via their Bayesian product:
\begin{equation}
    P_b=Norm({e^{-\alpha \cdot DT(M>\tau)}}\odot{\frac{Q(i,j) \cdot K(i,j)}{\gamma}} )
\end{equation}
A linear affine transform then decouples $P_b$ into spatially complementary retain weight R and enhance weight E:
\begin{equation}
    R=\mu_{R}(1-P_b)
\end{equation}
\begin{equation}
    E=\mu_{E} \cdot P_b
\end{equation}
where $\mu_{E}$,$\mu_{R}$,$\alpha$,$\gamma$ and $\tau$ are learnable scalar parameters. Higher R implies slower historical state decay; larger E implies stronger write weight for the current input.\par
\textbf{Differentiated Parameter Modulation in the Mamba Branch.} Standard Mamba imposes uniform $\Delta$ and B across all sequence positions—a reasonable assumption for natural images, but fundamentally inappropriate for medical segmentation where boundary and background pixels differ essentially in sequence modeling importance. Building on R and E, we apply position-differentiated modulation to $\Delta$ and B along the scan direction:
\begin{equation}
    \Delta_k=\Delta^{(k)}_0 \cdot (1-R_k)
\end{equation}
\begin{equation}
    B_k=B^{(k)}_0 \cdot (1+E_k)
\end{equation}
At boundary regions where $E_{k}→1$ and $R_{k}→0$,the spectral norm of $\bar{A}_k$ satisfies $\|\bar{A}_k\|_2 > \|\bar{A}_0\|_2$,retaining historical states more completely; simultaneously the Frobenius norm of $B_k$ increases, amplifying current input write weight—jointly imposing a local sequence memory gain achievable only at the architectural level.\par
\textbf{Proof:ablation validation is provided in Section~\ref{sec4.3}, Table~\ref{Table3}.}\par
\textbf{SE-Style Weight Fusion.} $F_{ssm}$, $F_{safs}$, and M are concatenated and passed through two independent prediction heads to generate spatial fusion weights $w_\text{ssm}$ and $w_\text{safs}$, regulated by a learnable temperature coefficient. The output is:
\begin{equation}
    F_{out}=w_{ssm} \cdot F_{ssm}+w_{safs} \cdot F_{safs}+x
\end{equation}
where x is the residual connection from the module input.\par
\textbf{Proposition 2.} The dual-branch design of BASM, combining boundary-posterior-driven Mamba parameter modulation and SASF local geometric compensation, forms a self-consistent internal mechanism that simultaneously achieves differentiated sequence memory at boundary regions and multi-scale structural inductive bias.\par
\textbf{Proof:ablation validation of the complete dual-branch design is provided in Section~\ref{sec4.3}, Table~\ref{Table3}. }
\subsection{Channel Mamba State Aggregation Module (CMSA)}\label{Sec3.3}
CMSA is deployed after the encoder bottleneck. Existing methods\cite{bib11,bib12,bib13} treat channels as independent parallel components, while SE-Net\cite{bib43} estimates only independent scalar weights, both failing to model inter-channel state dependencies\cite{bib44}. Motivated by Bau et al.\cite{bib42}, CMSA redefines the channel dimension as the SSM sequence dimension, learning inter-channel dependencies in a data-driven manner.\par
\textbf{Proposition 3 (Channel Sequence Non-Stationarity).} Let the feature map $\mathbf{X} \in \mathbb{R}^{C \times H \times W}$ treat the channel index as the sequence dimension. Since different channels respond to anatomically orthogonal concepts (texture, shape, density, etc.), adjacent channels satisfy distributional dissimilarity:
\begin{equation}
    \mu_k \not\approx \mu_{k+1}, \quad p(\mathbf{x}_k) \not\approx p(\mathbf{x}_{k+1})
\end{equation}
where $\mu_k$ denotes the spatial mean vector of the k-th channel and $p(\mathbf{x}_k)$ denotes its activation distribution across spatial positions. Under unconstrained standard SSM recurrence, the cumulative decay term may grow unboundedly as k increases:
\begin{equation}
    \prod_{i=1}^{k} \bar{A}_i=\prod_{i=1}^{k} e^{\Delta_iA}
\end{equation}
where $\bar{\mathbf{A}}_i$ is the ZOH-discretized state transition matrix and $\Delta_i$ the time-step of the i-th channel. Unbounded accumulation of heterogeneous channel states renders the recurrence theoretically non-convergent, necessitating an explicit boundedness constraint.\par
\textbf{Channel Serialization and Parameter Generation.} Flattening all C channels into a single sequence induces excessive heterogeneity, causing indiscriminate SSM state propagation. Inspired by grouped convolution, CMSA partitions channels into G groups of C/G semantically coherent channels, applying shared SSM parameters independently within each group to preserve inter-channel dependency modeling while bounding cross-group semantic contamination:
\begin{equation}
    h^{(g)}_k=f_{SSM}(x^{(g)}_k,h^{(g)}_{k-1}), \quad g=1,\dots,G
\end{equation}
where $g \in \{1, \ldots, G\}$ is the group index, $k \in \{1, \ldots, C/G\}$ the intra-group channel index,$\mathbf{x}_{k}^{(g)}$ the channel feature vector,$\mathbf{h}_{k}^{(g)}$ the SSM hidden state,$\mathbf{h}_{k-1}^{(g)}$ the preceding hidden state, and $f_{\text{SSM}}(\cdot)$ the bounded recurrence defined in Equations ~\ref{eq13} and ~\ref{eq14}. State propagation is strictly confined within each group.\par
\textbf{Bounded State Transition Constraint.} Unlike spatially adjacent positions that carry semantically continuous signals, adjacent channels may encode anatomically disparate concepts. Unconstrained SSM recurrence thus causes persistent inter-channel state contamination—unaddressed by existing methods \cite{bib11,bib12,bib13}. CMSA introduces a boundedness constraint on the state transition matrix and cumulative decay term:
\begin{equation}
    \bar{A}_k=clip(e^{\Delta_kA},0,\lambda)
    \label{eq13}
\end{equation}
\begin{equation}
     \prod_{i=1}^{k} \bar{A}_i=clip(\prod_{i=1}^{k} \bar{A}_i,0,\Lambda)
     \label{eq14}
\end{equation}
$\prod_{i=1}^{k} \bar{A}_i$ is the cumulative decay and $\Lambda$ the upper bound preventing state explosion, equivalent to imposing a Lipschitz restriction on cross-channel state propagation. Based on the bounded $\bar{\mathbf{A}}_k$,parallel approximate state aggregation is executed, and reconstructed features are residually connected to yield $\hat{\mathbf{X}}$.\par
\textbf{Proposition 4 (Boundedness Guarantee)}. Under the dual clipping constraints of Equations (13) and (14), the hidden state of any channel satisfies:
\begin{equation}
||h_k|| \leqslant \frac{||B_k|| \cdot ||u_k||}{1-\Lambda}
\end{equation}
$B_k$ and $\mathbf{u}_k$ are the input projection matrix and input vector of the k-th channel, and $\Lambda$ the learnable upper bound preventing state explosion—guaranteeing cross-channel semantic independence and distinguishing CMSA from naively channel-wise standard Mamba.\par
\textbf{Proof: Module effectiveness and hyperparameter robustness are validated in Section~\ref{sec4.3}, Table~\ref{Table3} and Section~\ref{Sec4.4}, Table~\ref{Table4}, respectively.}\par
\section{Experiments}\label{sec2}
\subsection{Datasets and Experimental Setup}\label{subsec2}
\textbf{Datasets.}We evaluate on two public benchmarks. The UW-Madison GI Tract~\cite{uwmadison2022} dataset comprises 16,481 2D MRI image–mask pairs from 85 patients across three hollow organ categories, split into 68 training (13,055 images) and 17 validation patients (3,426 images). The WORD\cite{bib29} dataset contains 150 3D CT cases with 16 organ categories, partitioned into 100 training and 20 validation cases.\par
\textbf{Experimental Setup.} All experiments use PyTorch on a Google Colab A100 GPU. Training employs AdamW $1 \times 10^{-4}$ , cosine annealing, batch size 4) with a Dice-CE combined loss augmented by deep supervision across five decoder outputs. Performance is reported in mDice and mIoU.
\subsection{Comparison with State-of-the-Art Methods}\label{subsec2}
We compare CS-MUNet against nine representative baselines on both datasets, with quantitative results summarized in Table~\ref{Table1}.\par
CS-MUNet achieves consistent state-of-the-art performance across both datasets, attaining mDice of 86.16\% on UW-Madison (+2.91\% over the strongest baseline) and 94.47\% on WORD (+1.03\%). BASM's boundary-posterior-driven modulation confers structural advantages in fine-grained boundary localization, while CMSA's sequential inter-channel modeling resolves cross-organ feature confusion—limitations that neither Transformer-based attention nor convolutional methods can overcome architecturally.
\begin{table*}[htbp]
  \centering
  \caption{Comparison with state-of-the-art methods on the UW-Madison GI Tract and WORD datasets.
           The best result in each group is shown in \textbf{bold}.}
  \label{Table1}
  \resizebox{\textwidth}{!}{%
  \setlength{\tabcolsep}{8pt}
  \renewcommand{\arraystretch}{0.92}
  \begin{tabular}{llccc}
    \toprule
    \textbf{Dataset} & \textbf{Method} & \textbf{mDice} & \textbf{mIoU} & \textbf{Params (M)} \\
    \midrule
    \multirow{10}{*}{\shortstack[l]{UW-Madison \\ GI Tract}}
      & U-Net~\cite{bib1}        & 0.8312 & 0.7562 & 32.66  \\
      & Att-UNet~\cite{bib21}                 & 0.8314 & 0.7564 & 67.85  \\
      & TransUNet~\cite{bib5}              & 0.8163 & 0.7394 & 117.19 \\
      & Swin-UNet~\cite{bib6}              & 0.8292 & 0.7531 & 41.48  \\
      & UNet++~\cite{bib2}                   & 0.8311 & 0.7487 & 92.63  \\
      & MSDUNet (2025)~\cite{bib22}          & 0.8325 & 0.7575 & 116.82 \\
      & SCUNet++ (2024)~\cite{bib23}        & 0.8325 & 0.7581 & 77.34  \\
      & EGE-UNet (2023)~\cite{bib24}         & 0.7819 & 0.7000 & 28.00  \\
      & nnU-Net~\cite{bib4}                  & 0.8245 & 0.7591 & 31.15  \\
      & \textbf{CS-MUNet (Ours)}               & \textbf{0.8616} & \textbf{0.7916} & 52.22 \\
    \midrule
    \multirow{10}{*}{WORD}
      & EGE-UNet (2023)~\cite{bib24}         & 0.8927 & 0.8695 & 28.00  \\
      & SCUNet++ (2024)~\cite{bib23}        & 0.8963 & 0.8736 & 77.34  \\
      & Swin-UNet~\cite{bib6}              & 0.9210 & 0.8972 & 41.48  \\
      & U-Net (Res2Net)~\cite{bib1} & 0.9228 & 0.9013 & 32.66 \\
      & TransUNet~\cite{bib5}              & 0.9287 & 0.9078 & 117.19 \\
      & MSDUNet (2025)~\cite{bib22}          & 0.9309 & 0.9092 & 116.82 \\
      & nnU-Net~\cite{bib4}                  & 0.9314 & 0.9130 & 31.15  \\
      & Att-UNet~\cite{bib21}                & 0.9340 & 0.9142 & 67.85  \\
      & UNet++~\cite{bib2}                   & 0.9344 & 0.9185 & 92.63  \\
      & \textbf{CS-MUNet (Ours)}               & \textbf{0.9447} & \textbf{0.9251} & 52.22 \\
    \bottomrule
  \end{tabular}%
  }
\end{table*}
\begin{table*}[htbp]
  \centering
  \caption{Per-organ segmentation results of CS-MUNet on the UW-Madison GI Tract and WORD datasets.
           HD95 is reported in mm; ASD is reported in mm.}
  \label{tab:per_organ}
  \setcounter{table}{2}
  \resizebox{\textwidth}{!}{%
  \setlength{\tabcolsep}{8pt}
  \renewcommand{\arraystretch}{1.00}  
  \begin{tabular}{llcccc}
    \toprule
    \textbf{Dataset} & \textbf{Organ} & \textbf{Dice} & \textbf{IoU} & \textbf{HD95 (mm)} & \textbf{ASD (mm)} \\
    \midrule
    \multirow{3}{*}{\shortstack[l]{UW-Madison \\ GI Tract}}
      & Stomach      & 0.8862 & 0.8421 &  6.83 & 2.35 \\
      & Small Bowel  & 0.8252 & 0.7400 & 15.41 & 3.87 \\
      & Large Bowel  & 0.8735 & 0.7927 & 15.19 & 4.03 \\
    \midrule
    \multirow{16}{*}{WORD}
      & Liver         & 0.9660 & 0.9512 &  5.34 & 1.40 \\
      & Right Kidney  & 0.9751 & 0.9639 &  2.38 & 0.85 \\
      & Spleen        & 0.9807 & 0.9699 &  1.73 & 0.64 \\
      & Pancreas      & 0.9809 & 0.9698 &  1.70 & 0.64 \\
      & Aorta         & 0.9449 & 0.9274 &  7.80 & 2.31 \\
      & IVC           & 0.9418 & 0.9303 &  4.71 & 1.68 \\
      & RAG           & 0.9368 & 0.9137 &  3.33 & 1.35 \\
      & LAG           & 0.9284 & 0.9097 &  9.53 & 3.02 \\
      & Gallbladder   & 0.8921 & 0.8650 & 13.08 & 4.10 \\
      & Esophagus     & 0.8608 & 0.8060 & 17.49 & 5.09 \\
      & Stomach       & 0.8953 & 0.8459 & 10.72 & 2.84 \\
      & Duodenum      & 0.9345 & 0.9191 & 11.71 & 4.27 \\
      & Left Kidney   & 0.9180 & 0.8966 &  3.91 & 1.75 \\
      & Tumor 1       & 0.9723 & 0.9621 &  3.76 & 1.37 \\
      & Tumor 2       & 0.9689 & 0.9574 &  1.72 & 0.78 \\
      & Tumor 3       & 0.9686 & 0.9584 &  1.66 & 0.70 \\
    \bottomrule
  \end{tabular}%
  }
\end{table*}

\subsection{Ablation Study}\label{sec4.3}
We design a systematic ablation study on both datasets, with results reported in Table~\ref{Table3}, using Res2Net-50 with a standard decoder as baseline and incrementally introducing each component.\par
\begin{table*}[htbp]
  \centering
  \caption{Ablation study of CS-MUNet on the UW-Madison GI Tract and WORD datasets.
           The best result in each group is shown in \textbf{bold}.}
  \label{Table3}
  \setlength{\tabcolsep}{10pt}
  \begin{tabular}{llcc}
    \toprule
    \textbf{Dataset} & \textbf{Configuration} & \textbf{mDice} & \textbf{mIoU} \\
    \midrule
    \multirow{6}{*}{\shortstack[l]{UW-Madison \\ GI Tract}}
      & Baseline                              & 0.8312 & 0.7562 \\
      & Baseline + BASM                       & 0.8432 & 0.7719 \\
      & Baseline + CMSA                       & 0.8484 & 0.7755 \\
      & w/o $\Delta$/B modulation             & 0.8537 & 0.7819 \\
      & w/o SE-style weight allocation        & 0.8484 & 0.7749 \\
      & \textbf{CS-MUNet (Ours)}              & \textbf{0.8616} & \textbf{0.7916} \\
    \midrule
    \multirow{6}{*}{WORD}
      & Baseline                              & 0.9228 & 0.9013 \\
      & Baseline + BASM                       & 0.9365 & 0.9162 \\
      & Baseline + CMSA                       & 0.9348 & 0.9151 \\
      & w/o $\Delta$/B modulation             & 0.9328 & 0.9130 \\
      & w/o SE-style weight allocation        & 0.9345 & 0.9140 \\
      & \textbf{CS-MUNet (Ours)}              & \textbf{0.9447} & \textbf{0.9251} \\
    \bottomrule
  \end{tabular}
\end{table*}
Introducing BASM and CMSA individually yields mDice gains of 1.20/1.37 and 1.72/1.20 points on UW-Madison/WORD; combining both further surpasses either in isolation, confirming complementary synergy.Removing $\Delta$/B modulation causes mDice drops of 0.79/1.19 points, demonstrating that boundary priors embedded in state transitions are indispensable; removing SE-style weight allocation degrades performance by over 1 point, validating that dynamic dual-stream fusion is necessary for spatially non-uniform boundary modeling.\par

\subsection{Hyperparameter Sensitivity Analysis}\label{Sec4.4}
To assess the robustness of the boundary modulation hyperparameters in BASM and the grouping number G in CMSA, we conduct a systematic sensitivity analysis on the UW-Madison dataset, with results reported in Table~\ref{Table4}.\par
\textbf{Sensitivity of $\mu_R$ and $\mu_E$.} The parameters $\mu_R$ and 
$\mu_E$ govern boundary state retention and input amplification respectively. 
mDice variation does not exceed 0.64 points across all configurations, with 
optimal $\mu_R{=}0.8$ and $\mu_E{=}1.2$. The low sensitivity confirms that 
performance gains arise from the architectural design of boundary-posterior-driven 
modulation rather than careful hyperparameter tuning.\par
\textbf{Sensitivity of CMSA Grouping Number $G$.} $G$ controls the trade-off 
between inter-channel dependency modeling capacity and cross-group semantic 
contamination. Performance monotonically declines as $G$ increases from 4 to 32, 
since larger groups truncate inter-channel dependency paths and degenerate toward 
SE-Net-style independent scalar reweighting. $G{=}4$ achieves the optimal balance 
between sequence length and parameter efficiency.
\begin{table*}[htbp]
  \centering
  \caption{Hyperparameter sensitivity analysis on the UW-Madison
           GI Tract dataset. All metrics are reported as
           percentages (\%). \textbf{Bold} indicates the optimal
           value in each group.}
  \label{Table4}
  \resizebox{\textwidth}{!}{%
  \renewcommand{\arraystretch}{0.92}
  \setlength{\tabcolsep}{18pt}
  \begin{tabular}{llccc}
    \toprule
    \textbf{Experiment} &
    \textbf{Value} &
    \textbf{mDice (\%)} &
    \textbf{mIoU (\%)} &
    \textbf{Optimal} \\
    \midrule
    \multirow{3}{*}{\shortstack[l]{Vary $\mu_{R}$ \\ (fixed $\mu_{E}\!=\!1.2$)}}
      & 0.5          & 85.64 & 78.58 & $\times$ \\
      & \textbf{0.8} & \textbf{86.16} & \textbf{79.16} & $\checkmark$ \\
      & 1.2          & 85.52 & 78.39 & $\times$ \\
    \midrule
    \multirow{3}{*}{\shortstack[l]{Vary $\mu_{E}$ \\ (fixed $\mu_{R}\!=\!0.8$)}}
      & 0.6          & 85.55 & 78.47 & $\times$ \\
      & \textbf{1.2} & \textbf{86.16} & \textbf{79.16} & $\checkmark$ \\
      & 1.5          & 85.69 & 78.60 & $\times$ \\
    \midrule
    \multirow{4}{*}{\shortstack[l]{Vary $G$ \\ (fixed $\mu_{R}\!=\!0.8$, \\ $\mu_{E}\!=\!1.2$)}}
      & \textbf{4}   & \textbf{86.16} & \textbf{79.16} & $\checkmark$ \\
      & 8            & 85.99 & 78.93 & $\times$ \\
      & 16           & 85.60 & 78.53 & $\times$ \\
      & 32           & 85.25 & 78.18 & $\times$ \\
    \bottomrule
  \end{tabular}%
  }
\end{table*}
\subsection{Parameter Efficiency}\label{subsec2}
\begin{figure*}[htbp]
    \centering
    \includegraphics[width=\textwidth,height=6cm]{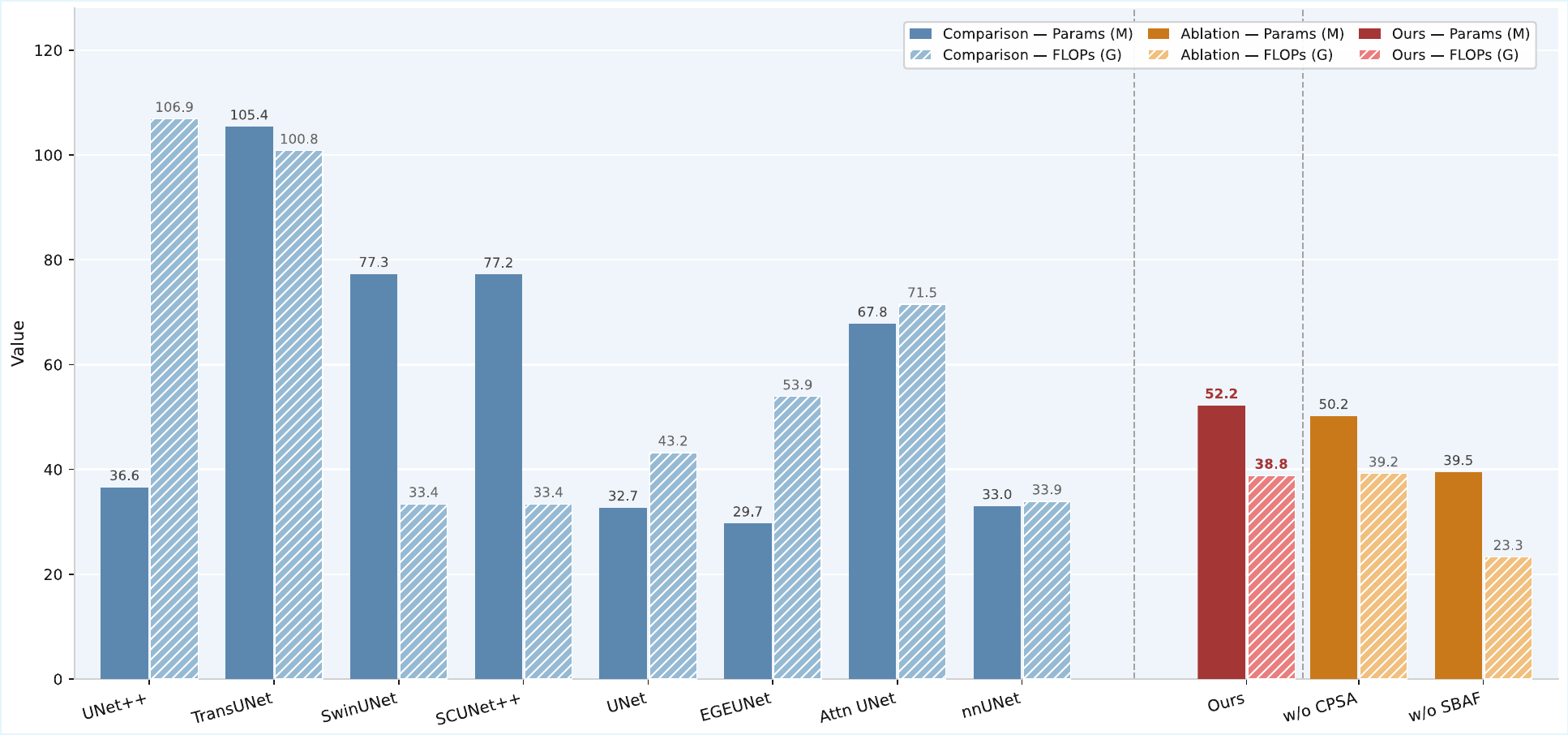}
   \caption{Comparison of parameter count (M) and computational cost (FLOPs, G) across all comparison models and ablation variants, with Ours highlighted in red.}
    \label{Figure 3}
\end{figure*}

To further assess the computational efficiency of the proposed method, we profile the parameter counts and floating-point operation (FLOPs) of all comparison models and ablation variants, with results shown in Figure~\ref{Figure 3}.\par
CS-MUNet achieves state-of-the-art performance with 52.2M parameters and 38.8G FLOPs—substantially fewer parameters than TransUNet and SwinUNet. Although CMSA and BASM introduce additional overhead over ablation variants, each cost is exchanged for substantive performance gains, demonstrating a favorable accuracy–efficiency trade-off with practical clinical deployment potential.
\subsection{Qualitative Visualization}\label{subsec2}
\begin{figure*}[!b]
    \centering
    \includegraphics[width=\textwidth,height=11cm]{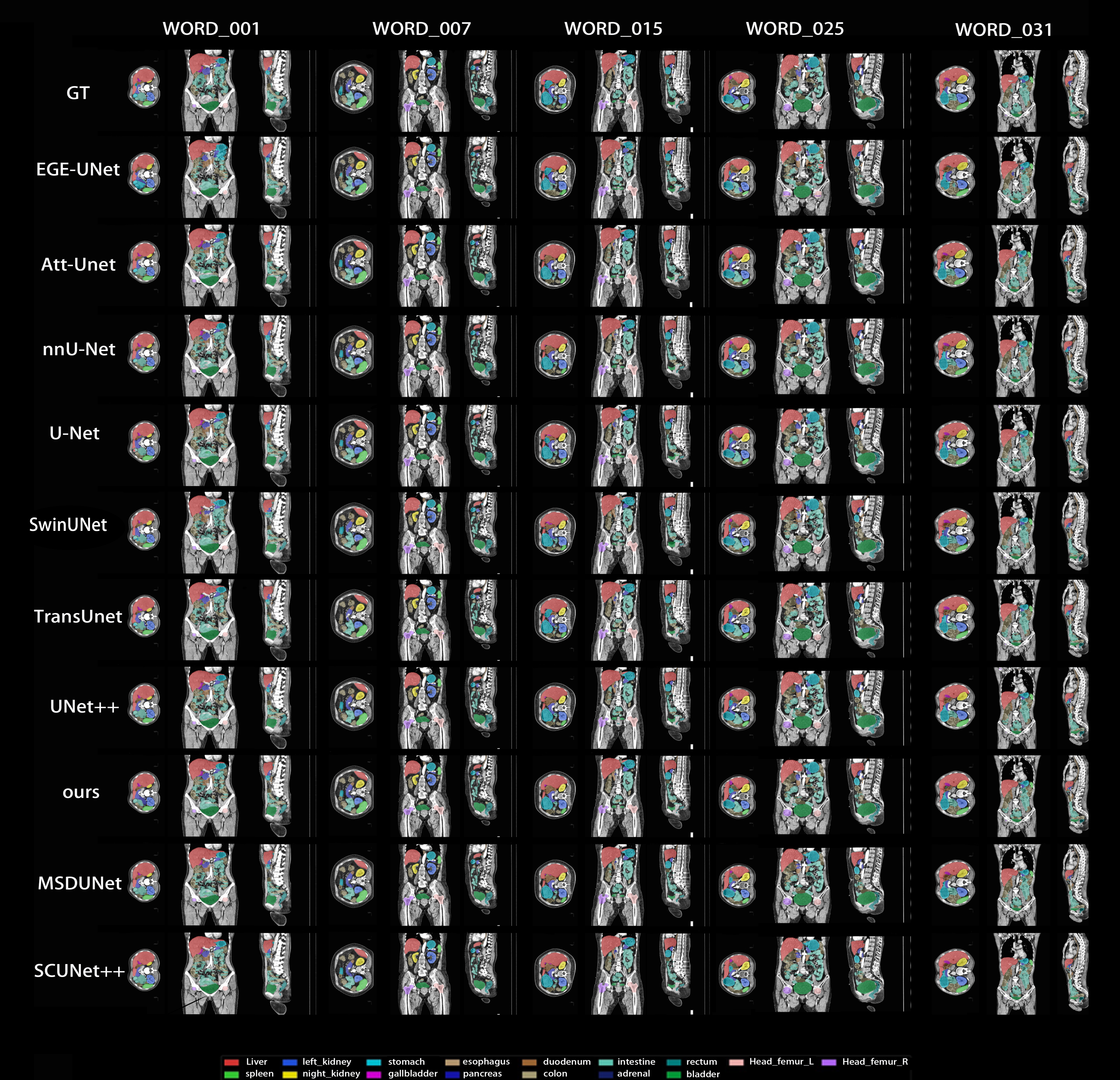}
   \caption{Qualitative segmentation comparisons across five representative WORD CT samples (coronal, sagittal, and axial views).}
    \label{Figure 4}
\end{figure*}

Figure~\ref{Figure 4} presents qualitative comparisons on five representative WORD samples across coronal, sagittal, and axial views. Most baseline methods produce reasonably complete segmentations for large-volume organs such as liver and spleen, but exhibit clear miss-detections, fragmentation, or boundary overflow on hollow organs such as small bowel and colon. The proposed method achieves the closest agreement with the ground truth, with notably superior contour continuity and boundary precision for intestinal organs.\par
\begin{figure*}[htbp]
    \centering
    \includegraphics[width=\textwidth,height=10cm]{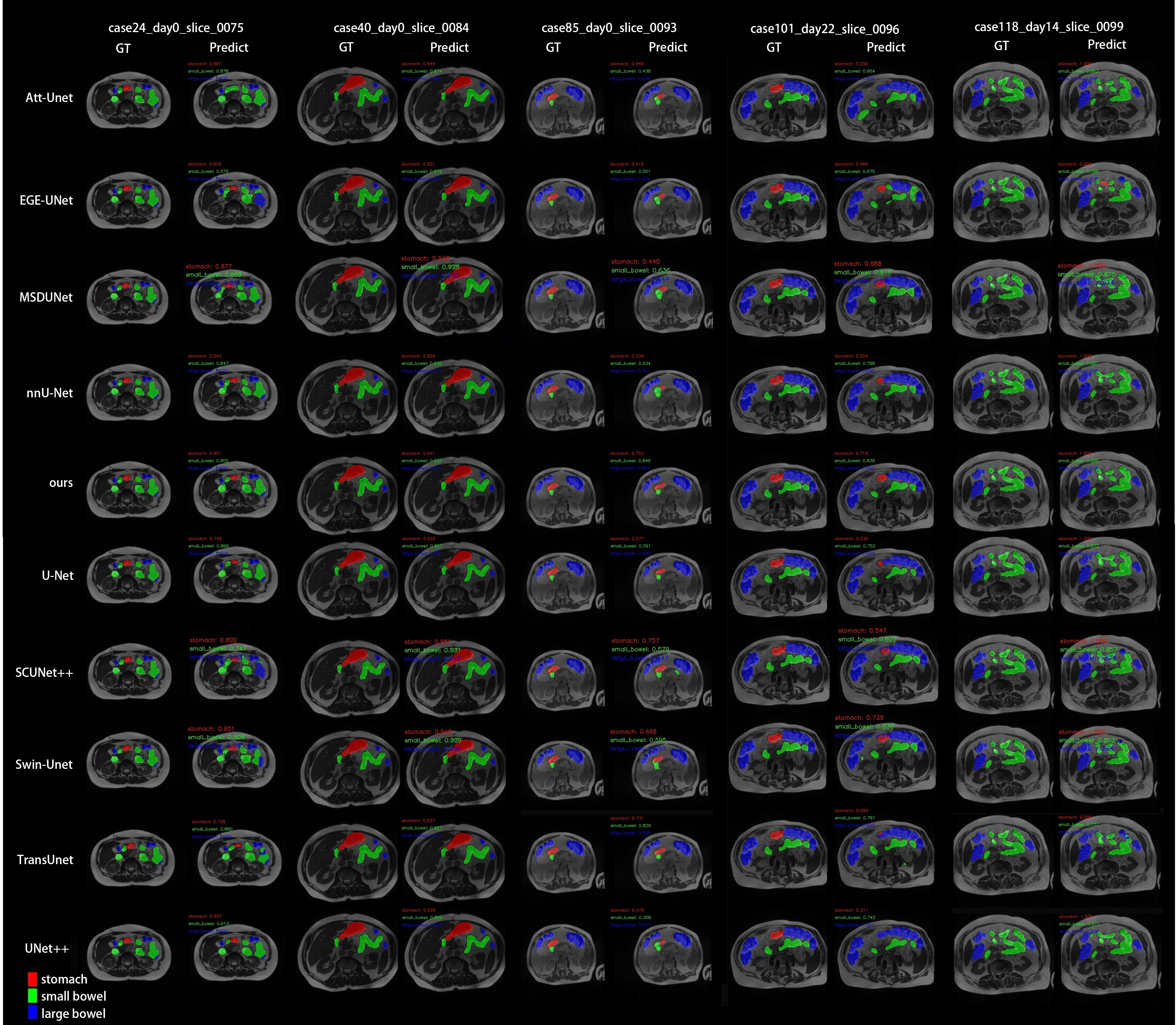}
   \caption{Per-slice segmentation comparisons on five representative UW-Madison MRI cases across three organ categories: stomach (red), small bowel (green), and large bowel (blue).}
    \label{Figure 5}
\end{figure*}

Figure~\ref{Figure 5} presents per-slice comparisons on the UW-Madison dataset across the three organ categories: stomach, small bowel, and large bowel. Most baselines exhibit regional fragmentation and boundary overflow in small bowel segmentation. The proposed method achieves high agreement with the ground truth across all three categories, with particularly outstanding contour continuity and boundary accuracy in small bowel segmentation.
\section{Discussion}\label{sec2}
CS-MUNet outperforms all nine baselines on both benchmarks. Ablation confirms that BASM's boundary-posterior-driven state transition modulation effectively mitigates fine-grained boundary localization deficiencies, while CMSA's sequential channel modeling mitigates the neglect of inter-channel anatomical semantic dependencies in existing spatial-only SSM designs\par
Existing methods scan exclusively along spatial dimensions without modeling inter-channel dependencies, while boundary approaches apply signals as auxiliary losses without modifying SSM parameters. In contrast, CMSA redefines channels as the SSM sequence dimension with explicit bounded constraints, and BASM injects boundary posteriors directly into $\Delta$ and B—embedding both capabilities at the architectural level rather than as external supervision.\par
Several limitations should be acknowledged. First, the 2D framework may limit volumetric boundary coherence for thin structures such as esophagus and duodenum; 3D extension is planned. Second, at 52.2M parameters, lightweight optimization remains unaddressed; distillation or pruning strategies are considered for future work. Third, generalization to ultrasound or PET-CT modalities requires further validation.\par
CS-MUNet demonstrates that jointly modeling spatial boundary semantics and inter-channel anatomical dependencies addresses two structural limitations of existing Mamba-based segmentation methods. Achieving mDice of 86.16\% and 94.47\% on UW-Madison and WORD respectively, with substantially fewer parameters than Transformer-based counterparts, CS-MUNet suggests practical deployment potential in clinical workflows requiring both segmentation accuracy and computational efficiency across MRI and CT modalities.
\section*{Funding}
This work was supported by the National Natural Science Foundation of China (No. 62562063),the Scientific Research Fund Project of the Yunnan and Provincial Department of Education (No. 2024J0024) and
the Youth Project of the National Social Science Fund of China (No. 25CFX009).
\section*{Declaration}
\textbf{Competing interests} The authors have no competing interests to de-clare that are relevant to the content of this article.
\section*{Data Availability}
The datasets analysed during the current study are publicly available.
The UW-Madison GI Tract dataset is available on Kaggle at
\url{https://www.kaggle.com/competitions/uw-madison-gi-tract-image-segmentation}\cite{uwmadison2022}
The WORD dataset is available at \url{https://github.com/HiLab-git/WORD}\cite{bib29}.

\begin{thebibliography}{45}
\ifx \bisbn   \undefined \def \bisbn  #1{ISBN #1}\fi
\ifx \binits  \undefined \def \binits#1{#1}\fi
\ifx \bauthor  \undefined \def \bauthor#1{#1}\fi
\ifx \batitle  \undefined \def \batitle#1{#1}\fi
\ifx \bjtitle  \undefined \def \bjtitle#1{#1}\fi
\ifx \bvolume  \undefined \def \bvolume#1{\textbf{#1}}\fi
\ifx \byear  \undefined \def \byear#1{#1}\fi
\ifx \bissue  \undefined \def \bissue#1{#1}\fi
\ifx \bfpage  \undefined \def \bfpage#1{#1}\fi
\ifx \blpage  \undefined \def \blpage #1{#1}\fi
\ifx \burl  \undefined \def \burl#1{\textsf{#1}}\fi
\ifx \doiurl  \undefined \def \doiurl#1{\url{https://doi.org/#1}}\fi
\ifx \betal  \undefined \def \betal{\textit{et al.}}\fi
\ifx \binstitute  \undefined \def \binstitute#1{#1}\fi
\ifx \binstitutionaled  \undefined \def \binstitutionaled#1{#1}\fi
\ifx \bctitle  \undefined \def \bctitle#1{#1}\fi
\ifx \beditor  \undefined \def \beditor#1{#1}\fi
\ifx \bpublisher  \undefined \def \bpublisher#1{#1}\fi
\ifx \bbtitle  \undefined \def \bbtitle#1{#1}\fi
\ifx \bedition  \undefined \def \bedition#1{#1}\fi
\ifx \bseriesno  \undefined \def \bseriesno#1{#1}\fi
\ifx \blocation  \undefined \def \blocation#1{#1}\fi
\ifx \bsertitle  \undefined \def \bsertitle#1{#1}\fi
\ifx \bsnm \undefined \def \bsnm#1{#1}\fi
\ifx \bsuffix \undefined \def \bsuffix#1{#1}\fi
\ifx \bparticle \undefined \def \bparticle#1{#1}\fi
\ifx \barticle \undefined \def \barticle#1{#1}\fi
\bibcommenthead
\ifx \bconfdate \undefined \def \bconfdate #1{#1}\fi
\ifx \botherref \undefined \def \botherref #1{#1}\fi
\ifx \url \undefined \def \url#1{\textsf{#1}}\fi
\ifx \bchapter \undefined \def \bchapter#1{#1}\fi
\ifx \bbook \undefined \def \bbook#1{#1}\fi
\ifx \bcomment \undefined \def \bcomment#1{#1}\fi
\ifx \oauthor \undefined \def \oauthor#1{#1}\fi
\ifx \citeauthoryear \undefined \def \citeauthoryear#1{#1}\fi
\ifx \endbibitem  \undefined \def \endbibitem {}\fi
\ifx \bconflocation  \undefined \def \bconflocation#1{#1}\fi
\ifx \arxivurl  \undefined \def \arxivurl#1{\textsf{#1}}\fi
\csname PreBibitemsHook\endcsname

\bibitem[\protect\citeauthoryear{Litjens et~al.}{2017}]{bib26}
\begin{barticle}
\bauthor{\bsnm{Litjens}, \binits{G.}},
\bauthor{\bsnm{Kooi}, \binits{T.}},
\bauthor{\bsnm{Bejnordi}, \binits{B.E.}},
\bauthor{\bsnm{al.}}:
\batitle{A survey on deep learning in medical image analysis}.
\bjtitle{Medical Image Analysis}
\bvolume{42},
\bfpage{60}--\blpage{88}
(\byear{2017})
\doiurl{10.1016/j.media.2017.07.005}
\end{barticle}
\endbibitem

\bibitem[\protect\citeauthoryear{Ma et~al.}{2022}]{bib27}
\begin{barticle}
\bauthor{\bsnm{Ma}, \binits{J.}},
\bauthor{\bsnm{Zhang}, \binits{Y.}},
\bauthor{\bsnm{Gu}, \binits{S.}},
\bauthor{\bsnm{al.}}:
\batitle{Abdomenct-1k: Is abdominal organ segmentation a solved problem?}
\bjtitle{IEEE Transactions on Pattern Analysis and Machine Intelligence}
\bvolume{44}(\bissue{10}),
\bfpage{6695}--\blpage{6714}
(\byear{2022})
\doiurl{10.1109/TPAMI.2021.3100536}
\end{barticle}
\endbibitem

\bibitem[\protect\citeauthoryear{Gibson et~al.}{2018}]{bib28}
\begin{barticle}
\bauthor{\bsnm{Gibson}, \binits{E.}},
\bauthor{\bsnm{Giganti}, \binits{F.}},
\bauthor{\bsnm{Hu}, \binits{Y.}},
\bauthor{\bsnm{al.}}:
\batitle{Automatic multi-organ segmentation on abdominal ct with dense
  v-networks}.
\bjtitle{IEEE Transactions on Medical Imaging}
\bvolume{37}(\bissue{8}),
\bfpage{1822}--\blpage{1834}
(\byear{2018})
\doiurl{10.1109/TMI.2018.2806309}
\end{barticle}
\endbibitem

\bibitem[\protect\citeauthoryear{Luo et~al.}{2022}]{bib29}
\begin{barticle}
\bauthor{\bsnm{Luo}, \binits{X.}},
\bauthor{\bsnm{Liao}, \binits{W.}},
\bauthor{\bsnm{Xiao}, \binits{J.}},
\bauthor{\bsnm{al.}}:
\batitle{Word: A large scale dataset, benchmark and clinical applicable study
  for abdominal organ segmentation from ct image}.
\bjtitle{Medical Image Analysis}
\bvolume{82},
\bfpage{102642}
(\byear{2022})
\doiurl{10.1016/j.media.2022.102642}
\end{barticle}
\endbibitem

\bibitem[\protect\citeauthoryear{Milletari et~al.}{2016}]{bib30}
\begin{bchapter}
\bauthor{\bsnm{Milletari}, \binits{F.}},
\bauthor{\bsnm{Navab}, \binits{N.}},
\bauthor{\bsnm{Ahmadi}, \binits{S.-A.}}:
\bctitle{V-net: Fully convolutional neural networks for volumetric medical
  image segmentation}.
In: \bbtitle{2016 Fourth International Conference on 3D Vision (3DV)},
pp. \bfpage{565}--\blpage{571}.
\bpublisher{IEEE},
\blocation{Stanford, CA, USA}
(\byear{2016}).
\doiurl{10.1109/3DV.2016.79}
\end{bchapter}
\endbibitem

\bibitem[\protect\citeauthoryear{Isensee et~al.}{2021}]{bib31}
\begin{barticle}
\bauthor{\bsnm{Isensee}, \binits{F.}},
\bauthor{\bsnm{Jaeger}, \binits{P.F.}},
\bauthor{\bsnm{Kohl}, \binits{S.A.A.}},
\bauthor{\bsnm{Petersen}, \binits{J.}},
\bauthor{\bsnm{Maier-Hein}, \binits{K.H.}}:
\batitle{nnu-net: a self-configuring method for deep learning-based biomedical
  image segmentation}.
\bjtitle{Nature Methods}
\bvolume{18}(\bissue{2}),
\bfpage{203}--\blpage{211}
(\byear{2021})
\doiurl{10.1038/s41592-020-01008-z}
\end{barticle}
\endbibitem

\bibitem[\protect\citeauthoryear{Ma et~al.}{2024}]{bib11}
\begin{botherref}
\oauthor{\bsnm{Ma}, \binits{J.}},
\oauthor{\bsnm{Li}, \binits{F.}},
\oauthor{\bsnm{Wang}, \binits{B.}}:
U-Mamba: Enhancing Long-range Dependency for Biomedical Image Segmentation.
Preprint at \url{https://arxiv.org/abs/2401.04722}
(2024)
\end{botherref}
\endbibitem

\bibitem[\protect\citeauthoryear{Xing et~al.}{2024}]{bib12}
\begin{bchapter}
\bauthor{\bsnm{Xing}, \binits{Z.}},
\bauthor{\bsnm{Ye}, \binits{T.}},
\bauthor{\bsnm{Yang}, \binits{Y.}},
\bauthor{\bsnm{Liu}, \binits{G.}},
\bauthor{\bsnm{Zhu}, \binits{L.}}:
\bctitle{Segmamba: Long-range sequential modeling mamba for 3d medical image
  segmentation}.
In: \bbtitle{Medical Image Computing and Computer Assisted Intervention --
  MICCAI 2024},
pp. \bfpage{578}--\blpage{588}.
\bpublisher{Springer},
\blocation{Cham}
(\byear{2024}).
\doiurl{10.1007/978-3-031-72111-3_54}
\end{bchapter}
\endbibitem

\bibitem[\protect\citeauthoryear{Ruan et~al.}{2025}]{bib13}
\begin{barticle}
\bauthor{\bsnm{Ruan}, \binits{J.}},
\bauthor{\bsnm{Li}, \binits{J.}},
\bauthor{\bsnm{Xiang}, \binits{S.}}:
\batitle{Vm-unet: Vision mamba unet for medical image segmentation}.
\bjtitle{ACM Transactions on Multimedia Computing, Communications, and
  Applications}
(\byear{2025})
\doiurl{10.1145/3767748}
\end{barticle}
\endbibitem

\bibitem[\protect\citeauthoryear{Ronneberger et~al.}{2015}]{bib1}
\begin{bchapter}
\bauthor{\bsnm{Ronneberger}, \binits{O.}},
\bauthor{\bsnm{Fischer}, \binits{P.}},
\bauthor{\bsnm{Brox}, \binits{T.}}:
\bctitle{U-net: Convolutional networks for biomedical image segmentation}.
In: \bbtitle{Medical Image Computing and Computer-Assisted Intervention --
  MICCAI 2015},
pp. \bfpage{234}--\blpage{241}.
\bpublisher{Springer},
\blocation{Cham}
(\byear{2015}).
\doiurl{10.1007/978-3-319-24574-4_28}
\end{bchapter}
\endbibitem

\bibitem[\protect\citeauthoryear{Zhou et~al.}{2018}]{bib2}
\begin{bchapter}
\bauthor{\bsnm{Zhou}, \binits{Z.}},
\bauthor{\bsnm{Siddiquee}, \binits{M.M.R.}},
\bauthor{\bsnm{Tajbakhsh}, \binits{N.}},
\bauthor{\bsnm{Liang}, \binits{J.}}:
\bctitle{Unet++: A nested u-net architecture for medical image segmentation}.
In: \bbtitle{Deep Learning in Medical Image Analysis and Multimodal Learning
  for Clinical Decision Support},
pp. \bfpage{3}--\blpage{11}.
\bpublisher{Springer},
\blocation{Cham}
(\byear{2018}).
\doiurl{10.1007/978-3-030-00889-5_1}
\end{bchapter}
\endbibitem

\bibitem[\protect\citeauthoryear{Huang et~al.}{2020}]{bib3}
\begin{bchapter}
\bauthor{\bsnm{Huang}, \binits{H.}},
\bauthor{\bsnm{Lin}, \binits{L.}},
\bauthor{\bsnm{Tong}, \binits{R.}},
\bauthor{\bsnm{al.}}:
\bctitle{Unet 3+: A full-scale connected unet for medical image segmentation}.
In: \bbtitle{ICASSP 2020 -- 2020 IEEE International Conference on Acoustics,
  Speech and Signal Processing},
pp. \bfpage{1055}--\blpage{1059}.
\bpublisher{IEEE},
\blocation{Barcelona, Spain}
(\byear{2020}).
\doiurl{10.1109/ICASSP40776.2020.9053405}
\end{bchapter}
\endbibitem

\bibitem[\protect\citeauthoryear{Isensee et~al.}{2021}]{bib4}
\begin{barticle}
\bauthor{\bsnm{Isensee}, \binits{F.}},
\bauthor{\bsnm{Jaeger}, \binits{P.F.}},
\bauthor{\bsnm{Kohl}, \binits{S.A.A.}},
\bauthor{\bsnm{Petersen}, \binits{J.}},
\bauthor{\bsnm{Maier-Hein}, \binits{K.H.}}:
\batitle{nnu-net: a self-configuring method for deep learning-based biomedical
  image segmentation}.
\bjtitle{Nature Methods}
\bvolume{18}(\bissue{2}),
\bfpage{203}--\blpage{211}
(\byear{2021})
\doiurl{10.1038/s41592-020-01008-z}
\end{barticle}
\endbibitem

\bibitem[\protect\citeauthoryear{Oktay et~al.}{2018}]{bib21}
\begin{botherref}
\oauthor{\bsnm{Oktay}, \binits{O.}},
\oauthor{\bsnm{Schlemper}, \binits{J.}},
\oauthor{\bsnm{Folgoc}, \binits{L.L.}},
\oauthor{\bsnm{al.}}:
Attention U-Net: Learning Where to Look for the Pancreas.
Preprint at \url{https://arxiv.org/abs/1804.03999}
(2018)
\end{botherref}
\endbibitem

\bibitem[\protect\citeauthoryear{Chen et~al.}{2024}]{bib5}
\begin{barticle}
\bauthor{\bsnm{Chen}, \binits{J.}},
\bauthor{\bsnm{Mei}, \binits{J.}},
\bauthor{\bsnm{Li}, \binits{X.}},
\bauthor{\bsnm{al.}}:
\batitle{Transunet: Rethinking the u-net architecture design for medical image
  segmentation through the lens of transformers}.
\bjtitle{Medical Image Analysis}
\bvolume{97},
\bfpage{103280}
(\byear{2024})
\doiurl{10.1016/j.media.2024.103280}
\end{barticle}
\endbibitem

\bibitem[\protect\citeauthoryear{Cao et~al.}{2023}]{bib6}
\begin{bchapter}
\bauthor{\bsnm{Cao}, \binits{H.}},
\bauthor{\bsnm{Wang}, \binits{Y.}},
\bauthor{\bsnm{Chen}, \binits{J.}},
\bauthor{\bsnm{al.}}:
\bctitle{Swin-unet: Unet-like pure transformer for medical image segmentation}.
In: \bbtitle{Computer Vision -- ECCV 2022 Workshops},
pp. \bfpage{205}--\blpage{218}.
\bpublisher{Springer},
\blocation{Cham}
(\byear{2023}).
\doiurl{10.1007/978-3-031-25066-8_9}
\end{bchapter}
\endbibitem

\bibitem[\protect\citeauthoryear{Hatamizadeh et~al.}{2022}]{bib7}
\begin{bchapter}
\bauthor{\bsnm{Hatamizadeh}, \binits{A.}},
\bauthor{\bsnm{Nath}, \binits{V.}},
\bauthor{\bsnm{Tang}, \binits{Y.}},
\bauthor{\bsnm{Yang}, \binits{D.}},
\bauthor{\bsnm{Roth}, \binits{H.R.}},
\bauthor{\bsnm{Xu}, \binits{D.}}:
\bctitle{Swin unetr: Swin transformers for semantic segmentation of brain
  tumors in mri images}.
In: \bbtitle{Brainlesion: Glioma, Multiple Sclerosis, Stroke and Traumatic
  Brain Injuries},
pp. \bfpage{272}--\blpage{284}.
\bpublisher{Springer},
\blocation{Cham}
(\byear{2022}).
\doiurl{10.1007/978-3-031-08999-2_22}
\end{bchapter}
\endbibitem

\bibitem[\protect\citeauthoryear{Gu and Dao}{2024}]{bib8}
\begin{bchapter}
\bauthor{\bsnm{Gu}, \binits{A.}},
\bauthor{\bsnm{Dao}, \binits{T.}}:
\bctitle{Mamba: Linear-time sequence modeling with selective state spaces}.
In: \bbtitle{First Conference on Language Modeling (COLM)}
(\byear{2024})
\end{bchapter}
\endbibitem

\bibitem[\protect\citeauthoryear{Zhu et~al.}{2024}]{bib9}
\begin{bchapter}
\bauthor{\bsnm{Zhu}, \binits{L.}},
\bauthor{\bsnm{Liao}, \binits{B.}},
\bauthor{\bsnm{Zhang}, \binits{Q.}},
\bauthor{\bsnm{Wang}, \binits{X.}},
\bauthor{\bsnm{Liu}, \binits{W.}},
\bauthor{\bsnm{Wang}, \binits{X.}}:
\bctitle{Vision mamba: Efficient visual representation learning with
  bidirectional state space model}.
In: \bbtitle{Proceedings of the 41st International Conference on Machine
  Learning (ICML 2024)},
pp. \bfpage{62393}--\blpage{62422}.
\bpublisher{PMLR},
\blocation{Vienna, Austria}
(\byear{2024})
\end{bchapter}
\endbibitem

\bibitem[\protect\citeauthoryear{Liu et~al.}{2024}]{bib10}
\begin{bchapter}
\bauthor{\bsnm{Liu}, \binits{Y.}},
\bauthor{\bsnm{Tian}, \binits{Y.}},
\bauthor{\bsnm{Zhao}, \binits{Y.}},
\bauthor{\bsnm{al.}}:
\bctitle{Vmamba: Visual state space model}.
In: \bbtitle{Advances in Neural Information Processing Systems 37 (NeurIPS
  2024)},
pp. \bfpage{103031}--\blpage{103063}
(\byear{2024})
\end{bchapter}
\endbibitem

\bibitem[\protect\citeauthoryear{Zhang et~al.}{2024}]{bib17}
\begin{bchapter}
\bauthor{\bsnm{Zhang}, \binits{M.}},
\bauthor{\bsnm{Yu}, \binits{Y.}},
\bauthor{\bsnm{Jin}, \binits{S.}},
\bauthor{\bsnm{Gu}, \binits{L.}},
\bauthor{\bsnm{Ling}, \binits{T.}},
\bauthor{\bsnm{Tao}, \binits{X.}}:
\bctitle{Vm-unet-v2: Rethinking vision mamba unet for medical image
  segmentation}.
In: \bbtitle{Bioinformatics Research and Applications: 20th International
  Symposium, ISBRA 2024},
pp. \bfpage{335}--\blpage{346}.
\bpublisher{Springer},
\blocation{Cham}
(\byear{2024}).
\doiurl{10.1007/978-981-97-5128-0_27}
\end{bchapter}
\endbibitem

\bibitem[\protect\citeauthoryear{Wang et~al.}{2024}]{bib32}
\begin{botherref}
\oauthor{\bsnm{Wang}, \binits{Z.}},
\oauthor{\bsnm{Zheng}, \binits{J.-Q.}},
\oauthor{\bsnm{Zhang}, \binits{Y.}},
\oauthor{\bsnm{Cui}, \binits{G.}},
\oauthor{\bsnm{Li}, \binits{L.}}:
Mamba-UNet: UNet-like pure visual Mamba for medical image segmentation.
Preprint at \url{https://arxiv.org/abs/2402.05079}
(2024)
\end{botherref}
\endbibitem

\bibitem[\protect\citeauthoryear{Gong et~al.}{2025}]{bib16}
\begin{bchapter}
\bauthor{\bsnm{Gong}, \binits{H.}},
\bauthor{\bsnm{Kang}, \binits{L.}},
\bauthor{\bsnm{Wang}, \binits{Y.}},
\bauthor{\bsnm{al.}}:
\bctitle{nnmamba: 3d biomedical image segmentation, classification and landmark
  detection with state space model}.
In: \bbtitle{2025 IEEE 22nd International Symposium on Biomedical Imaging
  (ISBI)}.
\bpublisher{IEEE},
\blocation{Houston, TX, USA}
(\byear{2025}).
\doiurl{10.1109/ISBI60581.2025.10980694}
\end{bchapter}
\endbibitem

\bibitem[\protect\citeauthoryear{Liu et~al.}{2024}]{bib14}
\begin{bchapter}
\bauthor{\bsnm{Liu}, \binits{J.}},
\bauthor{\bsnm{Yang}, \binits{H.}},
\bauthor{\bsnm{Zhou}, \binits{H.-Y.}},
\bauthor{\bsnm{al.}}:
\bctitle{Swin-umamba: Mamba-based unet with imagenet-based pretraining}.
In: \bbtitle{Medical Image Computing and Computer Assisted Intervention --
  MICCAI 2024},
pp. \bfpage{615}--\blpage{625}.
\bpublisher{Springer},
\blocation{Cham}
(\byear{2024}).
\doiurl{10.1007/978-3-031-72114-4_59}
\end{bchapter}
\endbibitem

\bibitem[\protect\citeauthoryear{Liao et~al.}{2024}]{bib15}
\begin{botherref}
\oauthor{\bsnm{Liao}, \binits{W.}},
\oauthor{\bsnm{Zhu}, \binits{Y.}},
\oauthor{\bsnm{Wang}, \binits{X.}},
\oauthor{\bsnm{Pan}, \binits{C.}},
\oauthor{\bsnm{Wang}, \binits{Y.}},
\oauthor{\bsnm{Ma}, \binits{L.}}:
LightM-UNet: Mamba Assists in Lightweight UNet for Medical Image Segmentation.
Preprint at \url{https://arxiv.org/abs/2403.05246}
(2024)
\end{botherref}
\endbibitem

\bibitem[\protect\citeauthoryear{Wu et~al.}{2025}]{bib33}
\begin{barticle}
\bauthor{\bsnm{Wu}, \binits{R.}},
\bauthor{\bsnm{Liu}, \binits{Y.}},
\bauthor{\bsnm{Ning}, \binits{G.}},
\bauthor{\bsnm{Liang}, \binits{P.}},
\bauthor{\bsnm{Chang}, \binits{Q.}}:
\batitle{Ultralight vm-unet: Parallel vision mamba significantly reduces
  parameters for skin lesion segmentation}.
\bjtitle{Patterns}
\bvolume{6}(\bissue{11}),
\bfpage{101298}
(\byear{2025})
\doiurl{10.1016/j.patter.2025.101298}
\end{barticle}
\endbibitem

\bibitem[\protect\citeauthoryear{Hao et~al.}{2024}]{bib18}
\begin{botherref}
\oauthor{\bsnm{Hao}, \binits{J.}},
\oauthor{\bsnm{Zhu}, \binits{Y.}},
\oauthor{\bsnm{He}, \binits{L.}},
\oauthor{\bsnm{Liu}, \binits{M.}},
\oauthor{\bsnm{Tsoi}, \binits{J.K.H.}},
\oauthor{\bsnm{Hung}, \binits{K.F.}}:
T-mamba: A unified framework with long-range dependency in dual-domain for 2d
  \& 3d tooth segmentation.
IEEE Transactions on Multimedia
\textbf{27}
(2024)
\doiurl{10.1109/TMM.2024.3405713}
\end{botherref}
\endbibitem

\bibitem[\protect\citeauthoryear{Wu et~al.}{2025}]{bib25}
\begin{barticle}
\bauthor{\bsnm{Wu}, \binits{R.}},
\bauthor{\bsnm{Liu}, \binits{Y.}},
\bauthor{\bsnm{Liang}, \binits{P.}},
\bauthor{\bsnm{Chang}, \binits{Q.}}:
\batitle{H-vmunet: High-order vision mamba unet for medical image
  segmentation}.
\bjtitle{Neurocomputing}
\bvolume{624},
\bfpage{129447}
(\byear{2025})
\doiurl{10.1016/j.neucom.2025.129447}
\end{barticle}
\endbibitem

\bibitem[\protect\citeauthoryear{Xu}{2024}]{bib34}
\begin{botherref}
\oauthor{\bsnm{Xu}, \binits{J.}}:
HC-Mamba: Vision MAMBA with Hybrid Convolutional Techniques for Medical Image
  Segmentation.
Preprint at \url{https://arxiv.org/abs/2405.05007}
(2024)
\end{botherref}
\endbibitem

\bibitem[\protect\citeauthoryear{Fan et~al.}{2025}]{bib35}
\begin{barticle}
\bauthor{\bsnm{Fan}, \binits{C.}},
\bauthor{\bsnm{Yu}, \binits{H.}},
\bauthor{\bsnm{Huang}, \binits{Y.}},
\bauthor{\bsnm{Wang}, \binits{L.}},
\bauthor{\bsnm{Yang}, \binits{Z.}},
\bauthor{\bsnm{Jia}, \binits{X.}}:
\batitle{Slicemamba with neural architecture search for medical image
  segmentation}.
\bjtitle{IEEE Journal of Biomedical and Health Informatics}
\bvolume{29}(\bissue{10}),
\bfpage{7446}--\blpage{7458}
(\byear{2025})
\doiurl{10.1109/JBHI.2025.3564381}
\end{barticle}
\endbibitem

\bibitem[\protect\citeauthoryear{Ma and Wang}{2024}]{bib36}
\begin{barticle}
\bauthor{\bsnm{Ma}, \binits{C.}},
\bauthor{\bsnm{Wang}, \binits{Z.}}:
\batitle{Semi-mamba-unet: Pixel-level contrastive and pixel-level
  cross-supervised visual mamba-based unet for semi-supervised medical image
  segmentation}.
\bjtitle{Knowledge-Based Systems}
\bvolume{300},
\bfpage{112203}
(\byear{2024})
\doiurl{10.1016/j.knosys.2024.112203}
\end{barticle}
\endbibitem

\bibitem[\protect\citeauthoryear{Wang and Ma}{2024}]{bib37}
\begin{botherref}
\oauthor{\bsnm{Wang}, \binits{Z.}},
\oauthor{\bsnm{Ma}, \binits{C.}}:
Weak-Mamba-UNet: Visual Mamba Makes CNN and ViT Work Better for Scribble-based
  Medical Image Segmentation.
Preprint at \url{https://arxiv.org/abs/2402.10887}
(2024)
\end{botherref}
\endbibitem

\bibitem[\protect\citeauthoryear{Xie et~al.}{2024}]{bib38}
\begin{botherref}
\oauthor{\bsnm{Xie}, \binits{J.}},
\oauthor{\bsnm{Liao}, \binits{R.}},
\oauthor{\bsnm{Zhang}, \binits{Z.}},
\oauthor{\bsnm{Yi}, \binits{S.}},
\oauthor{\bsnm{Zhu}, \binits{Y.}},
\oauthor{\bsnm{Luo}, \binits{G.}}:
ProMamba: Prompt-Mamba for polyp segmentation.
Preprint at \url{https://arxiv.org/abs/2403.13660}
(2024)
\end{botherref}
\endbibitem

\bibitem[\protect\citeauthoryear{Ye et~al.}{2024}]{bib39}
\begin{botherref}
\oauthor{\bsnm{Ye}, \binits{Z.}},
\oauthor{\bsnm{Chen}, \binits{T.}},
\oauthor{\bsnm{Wang}, \binits{F.}},
\oauthor{\bsnm{Zhang}, \binits{H.}},
\oauthor{\bsnm{Zhang}, \binits{L.}}:
P-Mamba: Marrying Perona Malik Diffusion with Mamba for Efficient Pediatric
  Echocardiographic Left Ventricular Segmentation.
Preprint at \url{https://arxiv.org/abs/2402.08506}
(2024)
\end{botherref}
\endbibitem

\bibitem[\protect\citeauthoryear{Yang et~al.}{2025}]{bib40}
\begin{barticle}
\bauthor{\bsnm{Yang}, \binits{Y.}},
\bauthor{\bsnm{Xing}, \binits{Z.}},
\bauthor{\bsnm{Yu}, \binits{L.}},
\bauthor{\bsnm{Huang}, \binits{C.}},
\bauthor{\bsnm{Fu}, \binits{H.}},
\bauthor{\bsnm{Zhu}, \binits{L.}}:
\batitle{Vivim: a video vision mamba for medical video segmentation}.
\bjtitle{IEEE Transactions on Circuits and Systems for Video Technology}
(\byear{2025})
\doiurl{10.1109/TCSVT.2025.3563411}
\end{barticle}
\endbibitem

\bibitem[\protect\citeauthoryear{Wang et~al.}{2022}]{bib19}
\begin{barticle}
\bauthor{\bsnm{Wang}, \binits{R.}},
\bauthor{\bsnm{Chen}, \binits{S.}},
\bauthor{\bsnm{Ji}, \binits{C.}},
\bauthor{\bsnm{Fan}, \binits{J.}},
\bauthor{\bsnm{Li}, \binits{Y.}}:
\batitle{Boundary-aware context neural network for medical image segmentation}.
\bjtitle{Medical Image Analysis}
\bvolume{78},
\bfpage{102395}
(\byear{2022})
\doiurl{10.1016/j.media.2022.102395}
\end{barticle}
\endbibitem

\bibitem[\protect\citeauthoryear{Sun et~al.}{2026}]{bib20}
\begin{barticle}
\bauthor{\bsnm{Sun}, \binits{L.}},
\bauthor{\bsnm{Duan}, \binits{P.}},
\bauthor{\bsnm{Li}, \binits{J.}}:
\batitle{Bamn: boundary-aware mamba network for skin lesion segmentation}.
\bjtitle{The Journal of Supercomputing}
\bvolume{82},
\bfpage{28}
(\byear{2026})
\doiurl{10.1007/s11227-025-08101-0}
\end{barticle}
\endbibitem

\bibitem[\protect\citeauthoryear{Li et~al.}{2025}]{bib22}
\begin{barticle}
\bauthor{\bsnm{Li}, \binits{X.}},
\bauthor{\bsnm{Li}, \binits{L.}},
\bauthor{\bsnm{Xing}, \binits{X.}},
\bauthor{\bsnm{al.}}:
\batitle{Msdunet: A model based on feature multi-scale and dual-input dynamic
  enhancement for skin lesion segmentation}.
\bjtitle{IEEE Transactions on Medical Imaging}
\bvolume{44}(\bissue{7}),
\bfpage{2894}--\blpage{2908}
(\byear{2025})
\doiurl{10.1109/TMI.2025.3549011}
\end{barticle}
\endbibitem

\bibitem[\protect\citeauthoryear{Chen et~al.}{2024}]{bib23}
\begin{bchapter}
\bauthor{\bsnm{Chen}, \binits{Y.}},
\bauthor{\bsnm{Zou}, \binits{B.}},
\bauthor{\bsnm{Guo}, \binits{Z.}},
\bauthor{\bsnm{al.}}:
\bctitle{Scunet++: Swin-unet and cnn bottleneck hybrid architecture with
  multi-fusion dense skip connection for pulmonary embolism ct image
  segmentation}.
In: \bbtitle{Proceedings of the IEEE/CVF Winter Conference on Applications of
  Computer Vision (WACV)},
pp. \bfpage{7759}--\blpage{7767}.
\bpublisher{IEEE},
\blocation{Waikoloa, HI, USA}
(\byear{2024}).
\doiurl{10.1109/WACV57701.2024.00758}
\end{bchapter}
\endbibitem

\bibitem[\protect\citeauthoryear{Ruan et~al.}{2023}]{bib24}
\begin{bchapter}
\bauthor{\bsnm{Ruan}, \binits{J.}},
\bauthor{\bsnm{Xie}, \binits{M.}},
\bauthor{\bsnm{Gao}, \binits{J.}},
\bauthor{\bsnm{Liu}, \binits{T.}},
\bauthor{\bsnm{Fu}, \binits{Y.}}:
\bctitle{Ege-unet: An efficient group enhanced unet for skin lesion
  segmentation}.
In: \bbtitle{Medical Image Computing and Computer Assisted Intervention --
  MICCAI 2023},
pp. \bfpage{481}--\blpage{490}.
\bpublisher{Springer},
\blocation{Cham}
(\byear{2023}).
\doiurl{10.1007/978-3-031-43901-8_46}
\end{bchapter}
\endbibitem

\bibitem[\protect\citeauthoryear{Lv et~al.}{2025}]{bib41}
\begin{barticle}
\bauthor{\bsnm{Lv}, \binits{C.}},
\bauthor{\bsnm{Li}, \binits{B.}},
\bauthor{\bsnm{Wang}, \binits{X.}},
\bauthor{\bsnm{al.}}:
\batitle{Ecm-transunet: Edge-enhanced multi-scale attention and convolutional
  mamba for medical image segmentation}.
\bjtitle{Biomedical Signal Processing and Control}
\bvolume{107},
\bfpage{107845}
(\byear{2025})
\doiurl{10.1016/j.bspc.2025.107845}
\end{barticle}
\endbibitem

\bibitem[\protect\citeauthoryear{Hu et~al.}{2018}]{bib43}
\begin{bchapter}
\bauthor{\bsnm{Hu}, \binits{J.}},
\bauthor{\bsnm{Shen}, \binits{L.}},
\bauthor{\bsnm{Sun}, \binits{G.}}:
\bctitle{Squeeze-and-excitation networks}.
In: \bbtitle{Proceedings of the IEEE Conference on Computer Vision and Pattern
  Recognition (CVPR)},
pp. \bfpage{7132}--\blpage{7141}.
\bpublisher{IEEE},
\blocation{Salt Lake City, UT, USA}
(\byear{2018}).
\doiurl{10.1109/CVPR.2018.00745}
\end{bchapter}
\endbibitem

\bibitem[\protect\citeauthoryear{Wang et~al.}{2020}]{bib44}
\begin{bchapter}
\bauthor{\bsnm{Wang}, \binits{Q.}},
\bauthor{\bsnm{Wu}, \binits{B.}},
\bauthor{\bsnm{Zhu}, \binits{P.}},
\bauthor{\bsnm{Li}, \binits{P.}},
\bauthor{\bsnm{Zuo}, \binits{W.}},
\bauthor{\bsnm{Hu}, \binits{Q.}}:
\bctitle{Eca-net: Efficient channel attention for deep convolutional neural
  networks}.
In: \bbtitle{Proceedings of the IEEE/CVF Conference on Computer Vision and
  Pattern Recognition (CVPR)},
pp. \bfpage{11534}--\blpage{11542}.
\bpublisher{IEEE},
\blocation{Seattle, WA, USA}
(\byear{2020}).
\doiurl{10.1109/CVPR42600.2020.01155}
\end{bchapter}
\endbibitem

\bibitem[\protect\citeauthoryear{Bau et~al.}{2017}]{bib42}
\begin{bchapter}
\bauthor{\bsnm{Bau}, \binits{D.}},
\bauthor{\bsnm{Zhou}, \binits{B.}},
\bauthor{\bsnm{Khosla}, \binits{A.}},
\bauthor{\bsnm{Oliva}, \binits{A.}},
\bauthor{\bsnm{Torralba}, \binits{A.}}:
\bctitle{Network dissection: Quantifying interpretability of deep visual
  representations}.
In: \bbtitle{Proceedings of the IEEE Conference on Computer Vision and Pattern
  Recognition (CVPR)},
pp. \bfpage{6541}--\blpage{6549}.
\bpublisher{IEEE},
\blocation{Honolulu, HI, USA}
(\byear{2017}).
\doiurl{10.1109/CVPR.2017.354}
\end{bchapter}
\endbibitem

\bibitem[\protect\citeauthoryear{{Kaggle and UW-Madison}}{2022}]{uwmadison2022}
\begin{botherref}
\oauthor{\bsnm{{Kaggle and UW-Madison}}}:
UW-Madison GI Tract Image Segmentation.
Kaggle
(2022).
\url{https://www.kaggle.com/competitions/uw-madison-gi-tract-image-segmentation}
\end{botherref}
\endbibitem

\end{thebibliography}

\end{document}